\pdfoutput=1

\documentclass[11pt]{article}

\usepackage[final]{acl}
\usepackage{booktabs} 
\usepackage{times}
\usepackage{latexsym}
\usepackage{amsmath}
\usepackage[T1]{fontenc}

\usepackage[utf8]{inputenc}

\usepackage{microtype}

\usepackage{inconsolata}
\usepackage{graphicx}

\title{Turbocharging Web Automation: The Impact of Compressed History States}

\author{Xiyue Zhu$^{\dagger}$\thanks{The work was done when Xiyue Zhu was an intern at Amazon. $^{**}$Peng Tang and Srikar Appalaraju are the corresponding authors.} \quad Peng Tang$^{\ddagger}$$^{**}$ \quad Haofu Liao$^{\ddagger}$ \quad Srikar Appalaraju$^{\ddagger}$$^{**}$\\
$^{\dagger}$University of Illinois at Urbana-Champaign \quad $^{\ddagger}$AWS AI Labs\\
\texttt{xiyuez2@illinois.edu, tangpeng723@gmail.com, \{liahaofu, srikara\}@amazon.com}}

\begin{document}
\maketitle
\begin{abstract}
Language models have led to a leap forward in web automation. The current web automation approaches take the current web state, history actions, and language instruction as inputs to predict the next action, overlooking the importance of history states.
However, the highly verbose nature of web page states can result in long input sequences and sparse information, hampering the effective utilization of history states.
In this paper, we propose a novel web history compressor approach to turbocharge web automation using history states.
Our approach employs a history compressor module that distills the most task-relevant information from each history state into a fixed-length short representation, mitigating the challenges posed by the highly verbose history states.
Experiments are conducted on the Mind2Web and WebLINX datasets to evaluate the effectiveness of our approach. Results show that our approach obtains 1.2-5.4\% absolute accuracy improvements compared to the baseline approach without history inputs.

\end{abstract}

\section{Introduction}

The task of web automation involves performing a sequence of actions to accomplish given tasks on any website, guided by language instructions \cite{deng2024mind2web,lu2024WebLINX,liu2018reinforcement,yao2022webshop,zhou2023webarena,park2025rvlm}.
Driven by advances in language models, web automation has attracted a lot of attention in recent years \cite{gur2023real,furuta2023multimodal,cheng2024seeclick,zheng2024gpt,park2025rvlm,Gao_2024_CVPR}.
These approaches leverage the current web state (i.e., web HTML and/or screenshot), history actions, and language instruction to predict the next action, obtaining promising web automation accuracy.

\begin{figure}
\includegraphics[width=\linewidth]{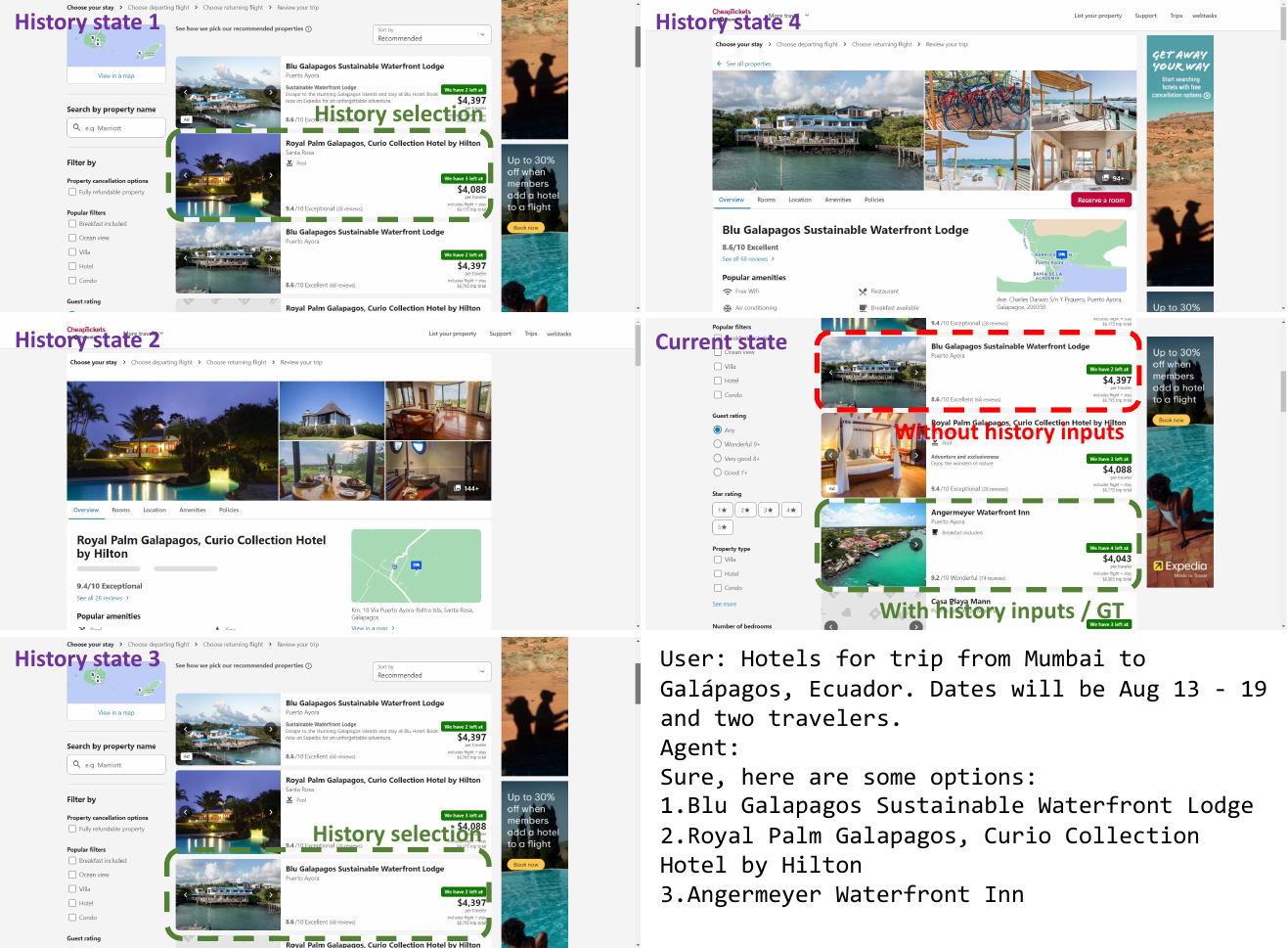}
\caption{Example results of w/ and w/o history inputs. Without seeing the histories, the model picks the same hotel in different steps. Adding history states in model inputs correctly picks different hotels in different steps.}

\label{fig:1}
\vspace{-0.4cm}
\end{figure}

However, the existing approaches do not consume history states, ignoring the fact that the history states are crucial to accomplish some web automation tasks, which leads to sub-optimal accuracy, see Figure \ref{fig:1}.

This fact motivates us to explore techniques that leverage history states to improve the accuracy of web automation.

The most straightforward way to leverage history states is to concatenate history states with other inputs (i.e., the current state, history actions, and language instruction) and feed the concatenated inputs into models.
But the state of real-world web pages could be very verbose \cite{deng2024mind2web,lu2024WebLINX}, posing several challenges to benefiting web automation models from the straightforward approach:
1) \textbf{Long input sequence.} Concatenating verbose history and current states results in a long input sequence, leading to high GPU memory cost and inference latency.
2) \textbf{Sparse information in history states.} Compared to the current state, the information that is relevant to the next action is much sparser in history states. Failing to effectively distilling the sparse information from history states could adversely impact the accuracy.
\begin{figure*}[!tb]
\centering
\includegraphics[width=\linewidth]{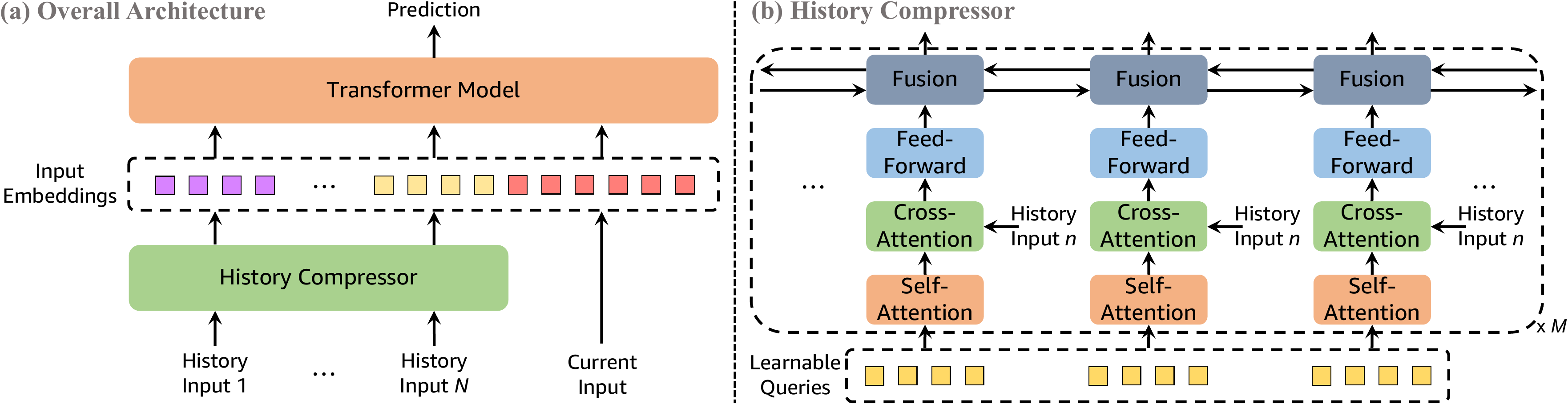}
\caption{
(a) Overall architecture of our model.
Our model takes the current input and $N$ history inputs as input, where the history inputs are fed into a history compressor before being fed into the transformer model. The next action is predicted based on the inputs.
(b) The architecture of the history compressor. For each history, the history compressor takes a fixed-length sequence of learnable queries and one history input as inputs with a history fusion module that fuses information among different history inputs, and outputs the representations of the learnable queries.
}
\label{fig:2}
\vspace{-0.4cm}
\end{figure*}

To address these challenges, we propose Web History Compressor, a novel approach to turbocharge web automation using history states.
Instead of feeding verbose history states into models directly, our approach trains a history compressor to compress each history state into a fixed-length short representation and extract the most relevant information.
Inspired by Perceiver \cite{Perceiver}, for each history state, the history compressor takes a fixed number of learnable queries, the history state, history actions, and language instructions as inputs, and outputs the representations of the learnable queries, effectively reducing the sequence length of the history state to the fixed number.
In addition, in the history compressor, the learnable queries cross-attend to the history state, history actions, and language instructions, {with information fusion among different history inputs,} allowing the history compressor to distill the most task-relevant information into the representations of the learnable queries, guided by the language instructions.
The resulting compressed representations of each history state are concatenated with other inputs (i.e., the current state, history actions, and language instruction). The concatenated inputs are fed into the model to predict the next action.

Experiments are carried out on the challenging Mind2Web \cite{deng2024mind2web}, and WebLINX \cite{lu2024WebLINX} datasets.
Our approach shows 1.2-5.4\% absolute accuracy improvements on the Mind2Web and WebLINX datasets across different evaluation metrics compared to the state-of-the-art MindAct approach \cite{deng2024mind2web}, confirming the effectiveness of our approach.

\vspace{-0.2cm}
\section{Approach}
\vspace{-0.1cm}
Figure \ref{fig:2}~(a) shows the architecture of our model.
Our model takes the current input and $N$ history inputs as input, where each input consists of state (i.e., web HTML and/or screenshot), history actions, and language instruction describing the task.
A history compressor module is employed to compress each history input into a fixed-length representation, see Section \ref{sec:history_compressor}.
Here, the history compressors for different history inputs share the same model weights.
The compressed representations of history inputs are concatenated with the current inputs, forming the input to a transformer model.
The transformer model then predicts the next action based on the inputs.
We apply a history compressor to history inputs only, because the current input is highly relevant to the task so we want to keep as much information in the current input as we can.

\vspace{-0.2cm}
\subsection{History Compressor}
\label{sec:history_compressor}
\vspace{-0.1cm}

The states of the real-world web pages could be highly verbose \cite{deng2024mind2web,lu2024WebLINX},
leading to long input sequences and sparse information in history states if we feed the history states into models directly.
Having an approach that can effectively distill the sparse information into compact representations can not only reduce the input sequence length but also improve model accuracy.
To address this, we propose a history compressor module to handle the verbose history state.

Inspired by Perceiver \cite{Perceiver}, each history compressor layer consists of a self-attention module, a cross-attention module, a feed-forward module, {and a history fusion module}.
Specifically, for each history input, a fixed number of learnable queries is first fed into the self-attention module,
and then input to a cross-attention module that cross-attends to one history input,
followed by a feed-forward module to enhance the learned representations.
{A history fusion module is next applied to fuse the information from the current history inputs and the neighboring history inputs.
The history fusion module operates by concatenating features from various history inputs along the channel dimension, followed by a fully-connected layer for dimension reduction.
The weights of learnable queries and different modules are shared across different history inputs.
See Figure \ref{fig:2}~(b).

With the history fusion module, different history inputs can communicate with each other to reduce the redundant information that are shared among different history inputs and learn the most useful information.}
With the guidance from the language instructions in the history input, the history compressor is able to distill the most task-relevant information into the learned representations.
By stacking $M$ history compressor layers, the learned representations of the fixed number of learnable queries are further enhanced for the task described in the language instruction.

\vspace{-0.2cm}
\subsection{Implementation Details}
\label{sec:implement}
\vspace{-0.1cm}

\noindent\textbf{Base Model} We use MindAct \cite{deng2024mind2web} as our base model.
Specifically, we use flan-T5-base \cite{raffel2020exploring,chung2024scaling} as our transformer model, which consists of encoder and decoder layers, and we use pruned HTML as the web state, following MindAct.
{Here the weights of History Compressor are randomly initialized.}

\noindent\textbf{Training} The representations of history inputs are derived from the outputs of the history compressor.
However, due to the absence of a history compressor for the current input, there is a misalignment between the different representations in the feature space.
This misalignment causes unstable training if we train all modules of our model jointly.
To mitigate this issue, we adopt a two-stage training strategy.
Specifically, in the first-stage training, we freeze the transformer model and train only the history compressor.
In addition, zero-initialized attention \cite{llamaadapter}, which assigns zero attention scores to history input representations at the beginning of training and gradually learns a factor that controls the attention weight assigned to these representations, is applied.
This approach allows the model to train the history compressor and align the representations of history inputs with those of the current input progressively, fostering a more stable and manageable training process.
Subsequently, in the second stage of training, we train all the modules together.

\vspace{-0.2cm}
\subsection{Alternative Approaches}
\label{sec:alter}
\vspace{-0.1cm}

There are alternative approaches for history compressors, including the pruning-based and zero-shot LLM compressors.
For the pruning-based compressor, we follow the prior work \cite{deng2024mind2web,lu2024WebLINX} by employing an off-the-shelf language model to rank and select {top-50 HTML elements}.
For the zero-shot LLM compressor, we carefully prompt strong LLMs (e.g., Claude-3 Haiku used here), with HTML, history actions, and language instruction as inputs, to summarize each history HTML. {In addition, we use previous work for LLM prompt compression, LLMLingua~\cite{jiang-etal-2023-llmlingua}, which uses pretrained LLMs to identify and remove non-essential tokens}. {Similar to our approach, $N$ history inputs are used for the two approaches.}
However, without task-specific training, these alternative approaches fail to distill the most task-relevant information into the compressed representations, resulting in unsatisfactory accuracy, see Section \ref{sec:results}.

\begin{table}[!tb]
\caption{Results of different approaches on the Mind2Web dataset. MindAct is the baseline approach without history inputs \cite{deng2024mind2web}. Pruning and LLM correspond to the alternative pruning-based compressor and zero-shot LLM compressor, respectively. {We also use LLMLingua \cite{jiang-etal-2023-llmlingua}, an existing work for LLM prompt compression. Ours indicates our history compressor approach. More details are in Section \ref{sec:alter}.} }
\vspace{-0.1cm}
\setlength{\tabcolsep}{3.5pt}
\centering
\resizebox{\linewidth}{!}{
\begin{tabular}{lcccc}
\toprule

  & Element  & Marco Element & Step  & Macro Step \\
  &  Acc ($\uparrow$) &  Acc ($\uparrow$) &  Acc ($\uparrow$) &  Acc ($\uparrow$) \\
\midrule 
\multicolumn{5}{c}{Cross-Task Split} \\

MindAct & 40.78 & 42.50 & 37.54 & 39.35 \\
Pruning & 38.54 & 41.39 & 36.15 & 38.81 \\
LLM & 34.53 & 39.07 & 32.04 & 36.37 \\
LLMLingua & 39.72&	42.37&	38.27&	39.23 \\
{Ours} & {\textbf{45.80}} & {\textbf{47.15}} & {\textbf{41.83}} & {\textbf{43.47}} \\
\midrule
\multicolumn{5}{c}{Cross-Website Split} \\%

MindAct & 29.57 & 31.96 & 26.37 & 28.54 \\
Pruning & 29.57 & 33.13 & 26.88 & 30.38 \\
LLM & 24.91 & 29.05 & 22.14 & 25.73 \\
LLMLingua & 30.27&	33.79&	23.38&	26.73 \\
{Ours} & {\textbf{32.17}} & {\textbf{35.71}} & {\textbf{28.73}} & {\textbf{31.83}} \\
\midrule
\multicolumn{5}{c}{Cross-Domain Split} \\

MindAct & 31.40 & 32.47 & 28.54 & 29.78 \\
Pruning & 31.79 & 33.54 & 29.00 & 30.94 \\
LLM & 29.27 & 29.28 & 26.70 & \textbf{31.80} \\
LLMLingua & 31.56&	32. 97&	28.79&	30.52 \\

{Ours} & {\textbf{32.65}} & {\textbf{33.90}} & {\textbf{29.70}} & {30.99} \\
\bottomrule

\label{tab:mind2web}
\end{tabular}
}
\vspace{-0.5cm}

\end{table}
\vspace{-0.2cm}
\section{Experiments}
\vspace{-0.1cm}

\subsection{Experimental Setup}
\vspace{-0.1cm}

\noindent\textbf{Datasets and Evaluation Metrics}
We benchmark our model on the challenging Mind2Web \cite{deng2024mind2web} and WebLINX \cite{lu2024WebLINX} datasets.
Following the official setups, we use element accuracy, macro element accuracy, step accuracy, and macro step accuracy as the evaluation metrics for Mind2Web, and overall micro average accuracy, overall intent-match accuracy, element-group IoU accuracy, and text-group F1 accuracy as the evaluation metrics for WebLINX.
We follow the official train/test splits for these two datasets.
Please see Appendix \ref{sec:supp_exp_data} for more details of datasets and evaluation metrics.

\noindent\textbf{Hyper-Parameter Setups} For each history compressor, 256 learnable queries with feature dimension 768 are used, and there are 2 history compressor layers. At most 5 history inputs are consumed by the model (i.e., $N=5$), see Appendix \ref{sec:supp_ablation_n_histories} for the impact of the number of history inputs.

\begin{table}[t]
\caption{Results of different approaches on the WebLINX test-iid dataset. MindAct is the baseline approach without history inputs \cite{deng2024mind2web}. PPruning and LLM correspond to the alternative pruning-based compressor and zero-shot LLM compressor, respectively, see Section \ref{sec:alter}. Ours indicates our history compressor approach.}
\vspace{-0.1cm}
\setlength{\tabcolsep}{3.5pt}
\centering
\resizebox{\linewidth}{!}{
\begin{tabular}{lcccc}
\toprule

  & Overall  & Overall  & Element-group  & Text-Group  \\
  &  Micro Avg ($\uparrow$) &  Intent-Match ($\uparrow$) &  IoU ($\uparrow$) &  F1 ($\uparrow$) \\
\midrule 
MindAct & 32.03 & 83.91 & 31.93 & 30.52 \\
Pruning & 31.37 & 83.32 & 30.65 & 28.61 \\
{LLM} & {31.24} & {82.90} & {31.29} & {29.01} \\

{Ours} & {\textbf{34.72}} & {\textbf{88.35}} & {\textbf{37.33}} & {\textbf{32.57}} \\
\bottomrule
\label{tab:WebLINX}
\end{tabular}
}
\vspace{-0.4cm}
\end{table}

\vspace{-0.2cm}
\subsection{Main Results}
\label{sec:results}
\vspace{-0.1cm}

The results presented in Table \ref{tab:mind2web} demonstrate the effectiveness of our proposed history compressor approach on the Mind2Web \cite{deng2024mind2web} dataset.
Across various evaluation metrics, our approach shows {1.2-5.0\%} accuracy improvement compared to the baseline approach of without history inputs.
This improvement confirms the ability of our approach to learn task-relevant representations for history inputs, thereby enhancing web automation performance.
In contrast, the alternative approaches, namely the pruning-based compressor and the zero-shot LLM compressor, yield mixed results.
While these approaches outperform the baseline on certain metrics, they also exhibit lower accuracy on others.
This is because, without task-specific training, these approaches cannot learn the most task-relevant information as the compressed representations, confirming the effectiveness of our approach.
Consequently, the superior performance of our approach highlights its effectiveness in leveraging history states for web automation tasks.
{In addition, different splits in Mind2Web evaluate the zero-shot generalization capabilities of the model (see Appendix \ref{sec:supp_exp_data}). Our approach consistently show improvements across different splits, confirming the generalization capabilities of our approach.}

The results on WebLINX \cite{lu2024WebLINX} shown in Tab.~\ref{tab:WebLINX} show the same conclusion as on Mind2Web.
In particular, our approach shows consistent performance gains over the baseline approach of without history inputs, with {2.0-5.4\%} accuracy improvements.
{Furthermore, our approach outperforms the pruning-based and zero-shot LLM compressor approaches by a substantial margin of 3.4-6.0\%.}
These results confirm the effectiveness of our approach.
The consistent improvements observed across multiple datasets and evaluation metrics provide compelling evidence of the robustness of our approach, further solidifying its effectiveness in web automation tasks.
\begin{table}[!tb]
\caption{Inference GPU memory cost and latency comparison among different approaches on WebLINX. MindAct is the baseline approach of without history inputs \cite{deng2024mind2web}. Pruning corresponds to the alternative pruning-based compressor described in Section \ref{sec:alter}. Ours indicates our history compressor approach. Experiments are performed on a single NVIDIA A100-SXM4-80GB GPU.}
\vspace{-0.1cm}
\setlength{\tabcolsep}{3.5pt}
\small\centering
\resizebox{\linewidth}{!}{
\begin{tabular}{lcccc}
\toprule
 & Inference GPU Memory & Inference Latency & {Average} & {Maximum} \\
 & (GB/sample $\downarrow$) & (s/demonstration $\downarrow$) & {\# tokens / history} & {\# tokens / history} \\
\midrule
{No compressor} & {-} & {-} & {$4065\pm318$} & {4096} \\
MindAct & \textbf{25.60} & \textbf{3.47} & {$\mathbf{0\pm0}$} & {\textbf{0}} \\
Pruning & 64.20 & 8.42 & {$1290\pm595$} & {2048} \\
{LLM} & {\underline{32.12}} & {7.25} & {\underline{$178\pm32$}} & {279} \\
{Ours} & {38.30} & {\underline{6.21}} & {$256\pm0$} & {\underline{256}} \\
\bottomrule
\label{tab:gpu_latency}
\end{tabular}
}
\vspace{-0.6cm}

\end{table}

\vspace{-0.2cm}
\subsection{Inference Cost Analysis}
\vspace{-0.1cm}

{We study the inference GPU memory cost, latency, and the number of input tokens here.
As shown in Tab.~\ref{tab:gpu_latency}, our approach obtained 40.3\% lower GPU memory cost, 26.2\% lower latency, and 5+ times fewer input tokens compared to the pruning-based compressor.}
This is because our approach effectively compresses the verbose history inputs into fixed-length learned representations, contributing to minimal inference overhead from including history inputs.
{Our approach has comparable GPU memory cost and latency compared to the zero-shot LLM compressor, with much higher web automation accuracy as shown in Tab.~\ref{tab:mind2web} and Tab.~\ref{tab:WebLINX}.}
We are not able to run training / inference for using history inputs without any compression due to the GPU memory limitation.

\vspace{-0.2cm}
\section{Conclusion}
\vspace{-0.1cm}
In this paper, we propose a novel web history compressor approach to improve web automation using history states.
Our approach trains a history compressor module to compress each highly verbose history state to a fixed-length short representation, maintaining the most task-relevant information meanwhile.
Experimental results show that our approach surpasses the baseline approach of without history inputs by 1.2-5.4\% on the Mind2Web and WebLINX datasets.
In the future, we will apply our approach to stronger transformer models and models that can take different modalities (e.g., image, image + text) as inputs, {and apply our approach to more web automation datasets}.

\section{Limitations}

Although our approach shows consistent improvements over the MindAct approach, theoretically, it can be applied to any base web automation models with any modality inputs, e.g., the stronger transformer model Llama 3 \cite{dubey2024llama} and the model with image modality inputs \cite{cheng2024seeclick} or multi-modality inputs \cite{dualview}.
Applying our approaches to these models could further confirm the effectiveness of our approach.
{In addition, although History Compressor gives significant accuracy boost for web automation, it also leads to 79\% higher latency compared to no history inputs. In the future, we will explore ways to reduce inference latency overhead, e.g., reducing the size of the history compressor model and reducing the number of learnable queries, etc.}
\bibliography{custom}

\newpage
\appendix

{
\section{Inputs and Outputs of the Transformer Model}
Here we provide details of the inputs and outputs of the transformer model in Figure \ref{fig:2}.
The transformer model used here is a Flan-T5 \cite{raffel2020exploring,chung2024scaling} based encoder-decoder model.
The inputs to the encoder layers are the concatenation of the representations of history inputs outputted from History Compressor, HTML representations of the current web state which are structured text inputs, and natural language instructions that describe the tasks to be accomplished. These inputs are processed through a series of encoder layers to produce encoded representations, which are then fed into the cross-attention layers of the decoder. The decoder performs autoregressive generation to produce the required output for task completion. 
}

\section{Datasets and Evaluation Metrics}
\label{sec:supp_exp_data}

{\noindent\textbf{Mind2Web} \cite{mind2web} has over 2,000 open-ended tasks from 137 websites and 31 domains.} There are 1,009 tasks from 73 websites for the training set. We evaluate our model on all 3 test splits: Cross-Task, Cross-Website, and Cross-Domain splits. {The Cross-Task split contains 252 tasks from 69 websites, and the training and testing sets contain websites from similar domains but different tasks to evaluate zero-shot task generalization. The Cross-Website split contains 10 websites and 177 tasks, and the training and testing sets contain websites from different domains but similar tasks to evaluate zero-shot domain generalization. The Cross-Domain split contains 912 tasks from 73 websites, and the training and testing sets contain websites from different domains and different tasks to evaluate zero-shot generalization across both website domains and task}. We evaluate our model using Element Accuracy and Step Success Rate. In the Mind2Web benchmark, the model is asked to select an element on the current HTML webpage and predict the action to perform on that element. Element Accuracy (element acc) compares the selected element with all acceptable elements, and the Step Success Rate (step acc) considers a step to be successful only if both the selected element and the predicted operation are correct. The raw metrics are calculated by averaging over each step. In contrast, the macro metrics average across tasks, each with a sequence of steps. Therefore, the macro metrics will weigh more on tasks consisting of short sequences of actions. 
\begin{table}[!tb]
\caption{Results of different numbers of history inputs of our approach on the Mind2Web dataset. 0 history means the MindAct baseline.}
\setlength{\tabcolsep}{2pt}
\centering
\resizebox{\linewidth}{!}{
{
\begin{tabular}{lcccc}
\toprule

  & Element  & Marco Element & Step  & Macro Step \\
  &  Acc ($\uparrow$) &  Acc ($\uparrow$) &  Acc ($\uparrow$) &  Acc ($\uparrow$) \\
\midrule 
\multicolumn{5}{c}{Cross-Task Split} \\

0 history & 40.78 & 42.50 & 37.54 & 39.35 \\
1 history & 42.26 & 42.90 & 38.12 & 40.89 \\
2 histories & 43.84 & 45.74 & 39.52 & 42.64 \\
3 histories & 43.36 & 43.64 & 38.92 & 41.78 \\
4 histories & 44.69 & 46.68 & 40.21 & 43.12\\
5 histories & \textbf{45.80} & \textbf{47.15} & \textbf{41.83} & \textbf{43.47} \\
\midrule
\multicolumn{5}{c}{Cross-Website Split} \\%
0 history   & 29.57  & 31.96 & 26.37 & 28.54 \\
1 history   & 29.08	& 31.43 & 25.96 & 28.18 \\
2 histories & 30.32	& 32.79 & 27.81 & 28.82 \\
3 histories & 31.21	& 33.89 & 28.13 & 31.43\\
4 histories & 32.01	& 35.60 & 28.36 & 31.52\\
5 histories & \textbf{32.17} & \textbf{35.71} & \textbf{28.73} & \textbf{31.83} \\
\midrule
\multicolumn{5}{c}{Cross-Domain Split} \\
0 history   &  31.40 & 32.48 & 28.54 & 29.78 \\
1 history   &  31.78 & 32.98 & 28.78 & 30.01 \\
2 histories &  31.98 & 33.13 & 29.45 & 30.33 \\
3 histories &  32.36 & 33.32 & 29.36 & 30.12 \\
4 histories &  32.45 & 33.68 & 29.62 & 30.51 \\
5 histories &  \textbf{32.65} & \textbf{33.90} & \textbf{29.70} & \textbf{30.99} \\
\bottomrule

\label{tab:ablation_n_histories}
\end{tabular}

}
}
\end{table}

\begin{table}[!tb]
\caption{{Results of different numbers of tokens used to represent each history input after compression on the Mind2Web dataset.}}
\setlength{\tabcolsep}{2pt}
\centering
\resizebox{\linewidth}{!}{
{
\begin{tabular}{lcccc}
\toprule

  & Element  & Marco Element & Step  & Macro Step \\
  &  Acc ($\uparrow$) &  Acc ($\uparrow$) &  Acc ($\uparrow$) &  Acc ($\uparrow$) \\
\midrule 
\multicolumn{5}{c}{Cross-Task Split} \\

64 tokens & 44.29 & 45.67 &	40.03 & 42.39 \\
128 tokens & 45.08&	46.36 &	41.07 &	42.97 \\
512 tokens & 45.49&	46.90 &	41.36 &	43.02 \\
256 tokens & \textbf{45.80} & \textbf{47.15} & \textbf{41.83} & \textbf{43.47} \\
\midrule
\multicolumn{5}{c}{Cross-Website Split} \\%
64 tokens   & 31.14	& 34.56 & 27.45 & 30.52 \\
128 tokens  & 31.70 & 35.01 & 28.34 & 31.42 \\
512 tokens  & 31.56 & 34.84 & 28.02 & 31.13 \\
256 tokens  & \textbf{32.17} & \textbf{35.71} & \textbf{28.73} & \textbf{31.83} \\
\midrule
\multicolumn{5}{c}{Cross-Domain Split} \\
64 tokens  &  31.98 & 33.14 & 29.39 & 30.48 \\
128 tokens &  32.47 & 33.76 & 29.58 & 30.74 \\
512 tokens &  32.01 & 33.24 & 29.23 & 30.32 \\
256 tokens &  \textbf{32.65} & \textbf{33.90} & \textbf{29.70} & \textbf{30.99} \\
\bottomrule

\label{tab:supp_ablation_histories_len}
\end{tabular}

}
}
\end{table}

{\noindent\textbf{WebLINX} \cite{weblinx} contains 2,337 demonstrations from 155 real-world websites.} There are 969 demonstrations with 43,538 turns in the training set. We evaluate our model on the Test\_iid split, containing 100 demos and 4,318 turns with similar tasks and websites in the training set to test the in-domain generalization of the model. The WebLINX benchmark requires text input from the model in \textbf{Text-Group} actions, such as load and say. It also contains \textbf{Element-Group} actions, such as click and submit, which require the model to select elements to perform the action. The metrics element group IoU and text group F1 calculate whether the text is correctly input and the elements are correctly selected, respectively. Intent match estimates whether the model can correctly predict which action to perform using accuracy. If an action is not correctly predicted in a certain step, the element group iou and text group F1 is 0 for this step. The overall micro average equals element group iou and text group F1, depending on the ground truth actions.

\begin{table}[!tb]
\caption{Results of without and with the history fusion module of our approach on the Mind2Web dataset. `With fusion` means our full approach.}
\setlength{\tabcolsep}{2pt}
\centering
\resizebox{\linewidth}{!}{
{
\begin{tabular}{lcccc}
\toprule

  & Element  & Marco Element & Step  & Macro Step \\
  &  Acc ($\uparrow$) &  Acc ($\uparrow$) &  Acc ($\uparrow$) &  Acc ($\uparrow$) \\
\midrule 
\multicolumn{5}{c}{Cross-Task Split} \\
Without fusion & 44.08 & 45.72 & 40.78 & 42.36 \\
With fusion & \textbf{45.80} & \textbf{47.15} & \textbf{41.83} & \textbf{43.47} \\
\midrule
\multicolumn{5}{c}{Cross-Website Split} \\%
Without fusion & \textbf{32.56}	& \textbf{36.15} & \textbf{28.84} & \textbf{31.95} \\
With fusion & 32.17 & 35.71 & 28.73 & 31.83 \\
\midrule
\multicolumn{5}{c}{Cross-Domain Split} \\
Without fusion &  32.47 & 33.58 & 29.59 & 30.84  \\
With fusion &  \textbf{32.65} & \textbf{33.90} & \textbf{29.70} & \textbf{30.99} \\
\bottomrule

\label{tab:ablation_fusion_mind2web}
\end{tabular}

}
}
\end{table}
\begin{table}[!tb]
\caption{Results of without and with the history fusion module of our approach on the WebLINX test-iid dataset. `With fusion` means our full approach.}
\setlength{\tabcolsep}{2pt}
\centering
\resizebox{\linewidth}{!}{
{
\begin{tabular}{lcccc}
\toprule

  & Overall  & Overall  & Element-group  & Text-Group  \\
  &  Micro Avg ($\uparrow$) &  Intent-Match ($\uparrow$) &  IoU ($\uparrow$) &  F1 ($\uparrow$) \\
\midrule 
Without fusion & 34.12 & 87.65 & 36.03 & 32.02 \\
With fusion & {\textbf{34.72}} & {\textbf{88.35}} & {\textbf{37.33}} & {\textbf{32.57}} \\
\bottomrule
\label{tab:ablation_fusion_weblinx}
\end{tabular}
}

}
\end{table}

{
\section{Ablation Studies}
\label{sec:supp_ablation}
Here we conduct experiments to analyze the impact of the number of history inputs, {the length of each compressed history state}, and the history fusion module.

\subsection{The Impact of the Number of History Inputs}
\label{sec:supp_ablation_n_histories}

We study the impact of the maximum number of histories on the Mind2Web dataset here.
The pruning-based compressor is used as the compression approach.
As shown in Tab.~\ref{tab:ablation_n_histories}, there is an overall trend that the more history inputs we incorporate, the better accuracy we can get.
The experiment results show that using history states can turbocharge web automation, which confirms our motivation that history states are important for web automation.
\subsection{The Impact of the Length of Each Compressed History State}
\label{sec:supp_ablation_histories_len}

{We investigate how the length of each compressed history state affects performance on the Mind2Web dataset. Our q-former-based compressor enables flexible control over the compression length by adjusting the number of learnable queries. As shown in Tab.~\ref{tab:supp_ablation_histories_len}, the optimal length of history states is 256. Using shorter compressed histories can lead to excessive information loss, resulting in sub-optimal performance. Conversely, longer compressed histories may dilute relevant information, making it harder for the web agent to utilize the past context effectively.}
\subsection{The Impact of the History Fusion Module}
Tab.~\ref{tab:ablation_fusion_mind2web} and Tab.~\ref{tab:ablation_fusion_weblinx} show the impact of the history fusion module on the Mind2Web and WebLINX datasets.
The integration of the history fusion module yields consistent improvements in web automation accuracy across the majority of test cases compared to without fusion. These results validate our hypothesis that the history fusion module effectively facilitates communication between different history inputs, enabling the model to extract and utilize the most relevant historical information for web automation.
}

\end{document}